\documentclass{article}

 \PassOptionsToPackage{numbers, compress}{natbib}

\usepackage[preprint]{neurips_2022}

\usepackage[most]{tcolorbox}
\usepackage{float}
\usepackage{xspace}
\tcbset{
  aibox/.style={
    width=474.18663pt,
    top=10pt,
    colback=white,
    colframe=black,
    colbacktitle=black,
    enhanced,
    center,
    attach boxed title to top left={yshift=-0.1in,xshift=0.15in},
    boxed title style={boxrule=0pt,colframe=white,},
  }
}
\newtcolorbox{AIbox}[2][]{aibox,title=#2,#1}

\usepackage[utf8]{inputenc} 
\usepackage[T1]{fontenc}    
\usepackage{hyperref}       
\usepackage{url}            
\usepackage{booktabs}       
\usepackage{amsfonts}       
\usepackage{nicefrac}       
\usepackage{microtype}      
\usepackage{xcolor}         
\usepackage{colortbl}

\usepackage{natbib}
\setcitestyle{numbers,square}

\usepackage[most]{tcolorbox}

\usepackage{listings}
\usepackage{makecell}
\usepackage{tabularray}

\usepackage{amsmath}
\usepackage{svg}
\usepackage{algorithm}
\usepackage{algorithmic}
\usepackage{framed}
\usepackage{adjustbox}
\usepackage{multirow}
\usepackage{graphicx}
\usepackage{amsthm}
\usepackage{amssymb}
\usepackage{enumerate}
\usepackage{enumitem}
\theoremstyle{plain}

\theoremstyle{definition}

\theoremstyle{remark}

\usepackage{pifont}
\newcommand{\cmark}{\ding{51}}%
\newcommand{\xmark}{\ding{55}}%

\definecolor{cblue}{RGB}{218,232,252}
\definecolor{cgreen}{RGB}{213,232,212}
\definecolor{cyellow}{RGB}{255,242,204}
\definecolor{cred}{RGB}{248,206,204}
\definecolor{corange}{RGB}{255,230,204}
\definecolor{cgray}{gray}{.9}

\usepackage{CJKutf8}
\usepackage{tabularx}

\makeatletter
\def\@fnsymbol#1{\ensuremath{\ifcase#1\or \includegraphics[width=0.45cm]{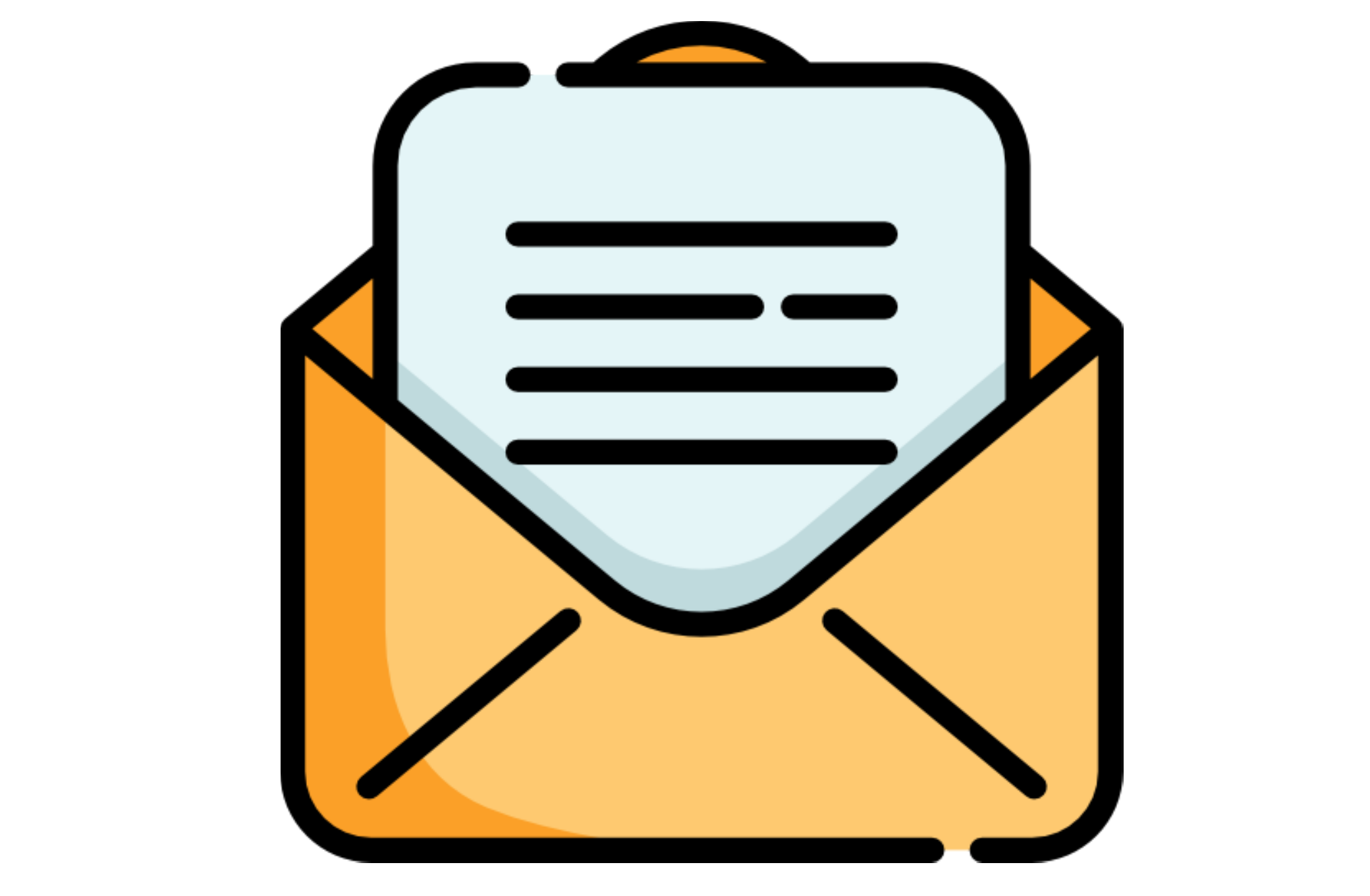} \or *\or 
   \mathsection\or \mathparagraph\or \|\or **\or \dagger\dagger
   \or \ddagger\ddagger \else\@ctrerr\fi}}
\makeatother

\title{HuatuoGPT, towards Taming Language Model\\ to Be a Doctor}  

\author{%
    Hongbo Zhang$^*$,
    Junying Chen$^*$,
    Feng Jiang$^*$,
    Fei Yu,
    Zhihong Chen,
    Jianquan Li,
    Guiming Chen \\
    \textbf{Xiangbo Wu},
    \textbf{Zhiyi Zhang},
    \textbf{Qingying Xiao},
    \textbf{Xiang Wan},
    \textbf{Benyou Wang}\thanks{Benyou is the corresponding author. The first three authors contributed to this work equally}~~~~, 
    \textbf{Haizhou Li}.\\    
   Shenzhen Research Institue of Big Data\\ 
    The Chinese University of Hong Kong, Shenzhen\\
  \texttt{wangbenyou@cuhk.edu.cn} \\
}

\begin{document}
\begin{CJK}{UTF8}{gkai}

\maketitle

\begin{abstract}


In this paper, we present HuatuoGPT, a large language model (LLM) for medical consultation. 
The  core recipe of HuatuoGPT is to leverage both \textit{distilled  data from ChatGPT} and \textit{real-world data from doctors}   in the supervised fine-tuned stage.  The responses of ChatGPT are usually detailed, well-presented and informative while it cannot perform like a doctor in many aspects, e.g. for  integrative diagnosis. We argue that real-world data from doctors would be complementary to distilled data in the sense the former could tame a distilled language model to perform like doctors. 
To better leverage the strengths of both data, we train a reward model to  align the language model with the merits that both data bring, following an RLAIF (reinforced learning from AI feedback) fashion.
To evaluate and benchmark the models, we propose a comprehensive evaluation scheme (including automatic and manual metrics). Experimental results demonstrate that HuatuoGPT  achieves state-of-the-art results in performing medical consultation among open-source LLMs in GPT-4 evaluation, human evaluation, and medical benchmark datasets. It is worth noting that by using additional real-world data and RLAIF, the distilled language model (i.e., HuatuoGPT) outperforms its teacher model ChatGPT in most cases. Our code, data, and models are publicly available at \url{https://github.com/FreedomIntelligence/HuatuoGPT}. The online demo is available at \url{https://www.HuatuoGPT.cn/}.

  
\end{abstract}

\section{Introduction}









\paragraph{LLMs for Medicine}
Medicine stands as a paramount pillar in human existence, encompassing profound significance. Medicine relies heavily on experiential knowledge, wherein seasoned physicians outperform their novice counterparts. However, the advent of generative artificial intelligence (AI) systems, such as ChatGPT and DALLE, which also can learn from past experiences and external sources, heralds a transformative era for experience-driven professions. It is increasingly evident that intelligent (or, say, `data-driven') medicine is an inexorable trend destined to materialize soon, albeit with ethical quandaries that demand consideration.


Medicine is a profoundly human endeavor where language plays a crucial role in facilitating interactions among clinicians, researchers, and patients. Coincidentally, the emergence of large language models (LLMs) in artificial intelligence is language-driven. This presents a remarkable opportunity for LLMs to contribute significantly to medicine. By bridging the gap between Medicine and LLMs,  referred to as LLM for Medicine or LLM4Med, large language models can bring about transformative changes in human lives. One such impact is the ability to provide equitable access to high-quality medical resources to people worldwide through online means. This aligns with the original vision of the internet era and fulfills the aspirations of AI.

It is distressing to envision the thousands of lives lost each day, particularly in underdeveloped areas, due to the unavailability of medical resources, untimely medical care, or exorbitant medical costs. Given the substantial disparities in medical resources across countries and even within a single country, LLMs for Medicine have the potential to address these imbalances and promote equality among all human beings.

\paragraph{Is ChatGPT Enough in Medicine?}
The short answer is `NO'. According to the recent study~\cite{ZBench}, it has been observed that ChatGPT, and even GPT-4, exhibit relatively poorer performance in vertical domains such as medicine. One contributing factor to this phenomenon is the potential lack of proficiency in medical knowledge among annotators. Consequently, there exist significant opportunities for further exploration and improvement in this domain.

On the other hand, online medicine often presents customized and localized challenges. For instance, Chinese medicine differs fundamentally from Western medicine, as does Indian medicine and many others. However, ChatGPT, being a general language model, lacks the capability for extensive customization. Additionally, entrusting private companies with users' medical data raises concerns, emphasizing the need for private deployment to ensure local data storage. Developing a  medical ChatGPT that is fully open-sourced and commercially viable would be advantageous for the well-being of individuals.



\paragraph{Use Cases and Advantages of LLM4Med}
The intended purposes of LLM4Med could be medical and health advice, triage, diagnosis, prescribing drugs, interpretation of medical reports, etc. In general, any medical or health information could be consolidated into an online chat process, similar to utilizing ChatGPT. Online medical consultation offers numerous advantages, including:

\begin{itemize}
\item Cost-effectiveness: The marginal cost of serving multiple users in an online manner is not linearly proportional to that of serving a single user. This scalability allows for cost-efficient expansion once the model is trained.
\item Reducing hospital crowding: The recent pandemic highlighted the risks associated with overcrowded hospitals, as many individuals sought offline consultations even when not requiring immediate medical treatment. By providing online alternatives, the strain on hospitals can be alleviated in order to mitigate the risks of future pandemics.
\item Addressing psychological barriers: Some individuals may refrain from seeking medical help or treatment due to fear or superstition, a phenomenon known as `讳病忌医' in Chinese. Online chatting platforms may provide a more comfortable environment for such individuals to discuss their concerns.
\end{itemize}

\paragraph{Our Practices for Chinese Medical LLM}
As widely recognized, healthcare inequality in China is a significant issue. Disparities in medical conditions between residents in first-tier cities and those in small cities and rural areas are striking. For instance, the average life expectancy in Shanghai stands at approximately 82 years, whereas in regions such as Guizhou, characterized by relative economic disadvantage, life expectancy drops significantly to 73 years.\footnote{\url{https://en.wikipedia.org/wiki/List_of_Chinese_administrative_divisions_by_life_expectancy}}

Here, we present a new  Chinese medical LLM called `HuatuoGPT' to commemorate the renowned Chinese physician Hua Tuo\footnote{\url{https://en.wikipedia.org/wiki/Hua_Tuo}}. Rather than training from real-world medical data as many previous language models did,  a straightforward way is to distill from ChatGPT ~\cite{vicuna2023,chen2023phoenix}  as it could quickly equip a language model with fluent chat and well-formatted responses.
However, distilling from ChatGPT in medical domain is problematic, since the teacher model (i.e. ChatGPT)  has the following issues:

\begin{itemize}
    \item ChatGPT does not perform well in medical domain, especially in Chinese~\cite{ZBench}.
    \item ChatGPT refuses to diagnose and prescribe drugs due to ethical and safety issues.
    \item ChatGPT does not perform as a doctor does. For example, it \textit{never ask} questions even though the patients' situation is incomplete for medical decision-making while doctors usually \textit{ask} for further details. In this case, ChatGPT gives a general response instead of a specialized one.
    \item ChatGPT struggles with hallucination due to the auto-regressive fashion.  
\end{itemize}

To overcome the above issues, the core recipe of HuatuoGPT is to leverage both \textit{real-world data from doctors} and \textit{distilled data from ChatGPT} in the Supervised Fine-Tuned (SFT) stage; both data consist of medical instruction data and medical conversation data~\cite{chen2023phoenix}. The distilled data from ChatGPT is used to tame language models to follow medical instructions and talk fluently. 
The additional  real-world medical data  not only inject medical knowledge into language models but also tame the language models to perform medical diagnoses or prescribe medications, act like a doctor and provide accurate information. The complementarity between real-world medical data and distilled data is further discussed in Sec.~\ref{sec:motivation}.

To leverage the strengths of both data (i.e., the real-world and distilled data) and meanwhile mitigate their weaknesses, we design a well-defined RL from AI Feedback (RLAIF)~\cite{bai2022constitutional} method after the SFT stage. It is used to reward the generated responses that are not only patient-friendly (learned from ChatGPT  with better presentation quality, lengthy and informative contents, instruction-following abilities and fluent chat), but also doctor-like (learned from doctors with professional and interactive diagnosis.). Technically, we employ LLMs to score generated responses based on their correctness, richness, logical consistency, and diagnostic ability to align our model with the both merits of ChatGPT and doctors. 

In assessing the performance of our model in the medical consultations, we meticulously crafted an evaluation schema encompassing both automated and manual assessments. HuatuoGPT, when assessed using GPT-4 in automatic evaluations on a series of 100 questions sourced from CBLUE with ten distinct medical intents, consistently outperformed incumbent Chinese medical models. More impressively, our model surpassed the performance of GPT-3.5-turbo in a majority of the evaluated cases. For the more complex multi-turn conversation evaluations, our HuatuoGPT model notably outshone ChatGPT in over 60\% of the instances in 20 departments, showcasing our proficiency in fusing real-world and distilled data and effectively applying reinforcement learning techniques to them. Furthermore, HuatuoGPT also achieved state-of-the-art (SOTA) performance in several medical benchmarks such as CmedQA, webmedQA, and Huatuo26M datasets.

To ensure the integrity and precision of our assessment, we incorporated manual evaluations of our model's performance in both single-turn and multi-turn conversation scenarios. The results from these manual evaluations corroborated the findings from our automated evaluations, thus reinforcing the reliability and consistency of our model's performance.


\paragraph{Significance of HuatuoGPT}

The contributions of HuatuoGPT are manyfold:

\begin{itemize}
    \item HuatuoGPT is the first medical language model to use RLAIF to leverage the merits of both real data and distilled data (including instruction and conversation data).
    \item This is among the first work that conducts systematic evaluation in medical LLMs.
    \item Human evaluation shows that  HuatuoGPT outperforms existing open-sourced LLMs and ChatGPT(GPT-3.5-turbo). Its performance is most similar to that of a doctor.
    \item We open-source our training data, code, HuatuoGPT model and the reward model at \url{https://github.com/FreedomIntelligence/HuatuoGPT}.

\end{itemize}





\section{Motivations}
\label{sec:motivation}

\begin{figure*}[t]
\small
\hspace{15pt}
\resizebox{0.95\textwidth}{!}{
\begin{AIbox}{ChatGPT vs.  Doctor}
{
\begin{tcolorbox}[colback=yellow!20, bottom=1pt, top=1pt, left=1pt, right=1pt]
{\bf User:} 我肚子痛 (\textit{I have a stomachache.})
\end{tcolorbox}
}
\tcbline
\begin{minipage}[t]{0.60\textwidth}
\vspace{-183.5pt}
\begin{tcolorbox}[colback=yellow!20, bottom=1pt, top=1pt, left=1pt, right=1pt] {\bf ChatGPT:} 我很抱歉听到你肚子痛。肚子痛可能有许多不同的原因，包括消化问题、感染、肠道问题或其他健康状况。我是一个语言模型，无法给出确切的医疗诊断，但我可以给你一些建议来缓解肚子痛的不适。(\textit{I'm sorry to hear that you have a stomachache. Stomachaches can have various causes, including digestive issues, infections, gastrointestinal problems, or other health conditions. As a language model, I cannot provide a precise medical diagnosis, but I can give you some advice to alleviate the discomfort of a stomachache.})

以下是一些可能有助于缓解肚子痛的常见方法 (Here are some common methods that may help alleviate a stomachache)：
\begin{itemize}[leftmargin=*]
\item 休息 (\textit{Rest})：给自己一些时间休息，放松身体，避免过度活动。(\textit{Give yourself some time to rest, relax your body, and avoid excessive physical activity.})
\item 热敷 (Warm compress)：可以尝试在肚子疼痛的区域放置热水袋或热毛巾，这有时可以缓解不适。(You can try placing a hot water bottle or warm towel on the area of the stomachache. This can sometimes help alleviate discomfort.)
\item 药物 (\textit{Medication})：可以尝试一些非处方的药物，例如抗酸药、轻度止痛药或抗胀气药。然而，在使用任何药物之前，最好先咨询医生或药剂师的建议。(\textit{You can try over-the-counter medications such as antacids, mild pain relievers, or anti-gas medications. However, it's advisable to consult a doctor or pharmacist for advice before using any medication.})
\item 医疗咨询 (\textit{Medical consultation})：如果肚子痛持续或加重，或伴随其他严重症状（如持续呕吐、高烧、血便等），请尽快咨询医生，以获取专业的建议和治疗。(\textit{If your stomachache persists or worsens, or if it is accompanied by other severe symptoms such as persistent vomiting, high fever, bloody stools, etc., please consult a doctor as soon as possible for professional advice and treatment.})
\end{itemize}

记住，以上建议仅供参考，如果你的症状持续存在或加重，最好咨询医生以获取专业的诊断和治疗建议。(\textit{Please remember that the above suggestions are for reference only. If your symptoms persist or worsen, it is best to consult a doctor for a professional diagnosis and treatment advice.})
\end{tcolorbox}
\end{minipage}
\begin{minipage}{0.395\textwidth}
\begin{tcolorbox}[colback=yellow!20, bottom=1pt, top=1pt, left=1pt, right=1pt]
  {\bf Doctor:}  { \color{blue} 您的腹痛是在哪里？是剧烈的还是轻微的？持续多久了？(\textit{Where is your abdominal pain? Is it severe or mild? How long has it been going on?})}
\end{tcolorbox}
\begin{tcolorbox}[colback=yellow!20, bottom=1pt, top=1pt, left=1pt, right=1pt]
  {\bf User:}  下腹部，有点疼，昨晚开始的，有小半天了。(\textit{Lower abdomen, a bit painful, started last night, been about half a day.})
\end{tcolorbox}
\begin{tcolorbox}[colback=yellow!20, bottom=1pt, top=1pt, left=1pt, right=1pt]
  {\bf Doctor:} { \color{blue} 是否伴随其他症状如恶心、呕吐或发热等？(\textit{Are there any accompanying symptoms such as nausea, vomiting, or fever?})}
\end{tcolorbox}
\begin{tcolorbox}[colback=yellow!20, bottom=1pt, top=1pt, left=1pt, right=1pt]
  {\bf User:}  有点呕吐。(\textit{I vomited.})
\end{tcolorbox}
\begin{tcolorbox}[colback=yellow!20, bottom=1pt, top=1pt, left=1pt, right=1pt]
  {\bf Doctor:}  {\color{blue} 昨晚吃什么了？(\textit{What did you eat last night?})}
\end{tcolorbox}
\begin{tcolorbox}[colback=yellow!20, bottom=1pt, top=1pt, left=1pt, right=1pt]
  {\bf User:}  昨晚吃了点小龙虾。会不会是这个导致的？(\textit{Eating some crayfish last night, could that be the cause?})
\end{tcolorbox}
\begin{tcolorbox}[colback=yellow!20, bottom=1pt, top=1pt, left=1pt, right=1pt]
  {\bf Doctor:}  很可能是\underline{急性肠胃炎}，通常是由于食物中毒引起的，而螺蛳粉中含有细菌和寄生虫，很容易引起肠胃感染。(\textit{It's highly possible that you have \underline{acute gastroenteritis}, which is often caused by food poisoning. Snail noodles, in particular, can contain bacteria and parasites that easily lead to gastrointestinal infections.})
\end{tcolorbox}
\begin{tcolorbox}[colback=yellow!20, bottom=1pt, top=1pt, left=1pt, right=1pt]
  {\bf User:}  ...
\end{tcolorbox}
\end{minipage}
\end{AIbox}}
\caption{Example of ChatGPT responses (left) and doctor responses (right) in  Medical Consultation Dialogue, where texts are \textit{translated} from Chinese to English. { \color{blue}  Questions raised by doctors 
}  are in { \color{blue}  blue}, and \underline{medical diagnoses} are underlined. Note that \textbf{ChatGPT usually does not 
raise questions in response to patients or provide medical diagnoses like doctors}.}
\label{fig:Example of Real-world (left) and Synthetic (right) Medical Consultation Dialogue}
\end{figure*}

\subsection{Learning from Humans or ChatGPT?}
Training language models from purely real-world conversation was a common practice~\cite{wei2018task,xu2019end,xia2020generative,zeng2020meddialog}. However, this suffers from low-quality data. For example, the responses in real-world conversations might be uninformative, short, and poorly presented. More importantly, the values in these data are not aligned and even contradictory. Learning from purely humans usually result in an unsatisfied chat-based language model~\cite{zhang2019dialogpt,thoppilan2022lamda,mi2022pangubot} compared to ChatGPT.

Recent work tends to distill a language model from ChatGPT, either imitating ChatGPT responses from single-turn instructions~\cite{alpaca} or learning the ChatGPT responses when interactively chatting with humans~\cite{vicuna2023}.   By distilling output from ChatGPT, a model can quickly acquire impressive instruction-following capabilities and seamless dialogue skills. In addtion, characterized by its diversity and rapid generation,  ChatGPT-distilled data can span various medical dialogues, encompassing various diseases, symptoms, and treatment modalities. This breadth and diversity substantially enhance the predictive performance and generalizability of the model. 



\subsection{Learning From Both Doctors and Chatgpt in Medicine}

\paragraph{Complementarity Between ChatGPT and Doctors}
However, distillation from ChatGPT might not work for medical LLMs since there exists a fundamental  gap between ChatGPT responses and  doctor responses, as shown in Figure~\ref{fig:Example of Real-world (left) and Synthetic (right) Medical Consultation Dialogue} and Table~\ref{tab:comparision}. The quality of distilled data can fluctuate, manifesting as incorrect or ambiguous information in the generated conversations. 
Contrastingly, real-world data, harvested from authentic doctor-patient interactions, provide an indispensable perspective into the complexities of actual medical scenarios. It can accurately reflect the true intention distribution of patients and has accurate diagnoses from doctors. The primary strength of real-world data lies in its high accuracy and professionalism. 

\begin{table}[h]
\small
    \centering
    \resizebox{\textwidth}{!}{
    \begin{tabular}{llcc|ll}
        \toprule
            & \multirow{2}{*}{Aspects }  &  \multicolumn{2}{|c|}{Responses}  &   \multirow{2}{*}{How ChatGPT works  }\\
            && \multicolumn{1}{|l} {ChatGPT}  & \multicolumn{1}{l|} {Doctor}  &   \\
        \midrule
        & \multirow{5}{*}{diagnostic ability}  &  \multirow{5}{*}{\xmark}  &  \multirow{5}{*}{\cmark}  &  \begin{varwidth}{0.6\textwidth}     {\bf Example:} \end{varwidth}  \\ &&&&
        \begin{varwidth}{0.6\textwidth}    {\bf User:} 医生，我肚子痛是得了肠胃炎吗？(Doctor, do I have gastroenteritis as I have a stomachache?)  \end{varwidth}  \\
        &&&& \begin{varwidth}{0.6\textwidth}   \textbf{ChatGPT}: 作为一个语言模型，我不能进行医学诊断 (As a language model, I am unable to provide medical diagnoses.)\end{varwidth}  \\
        \cline{2-5}
         \multirow{3}{*}{doctor-like}  & \multirow{3}{*}{expert-level accuracy} 
          &  \multirow{3}{*}{low}   &  \multirow{3}{*}{ \textbf{high} }  &  \begin{varwidth}{0.7\textwidth} \textbf{Example}: \end{varwidth}   \\
          &&&&
          \begin{varwidth}{0.7\textwidth} \textbf{User}: 什么是`勾三股四弦五'?  \end{varwidth}   \\
          &&&&  \begin{varwidth}{0.7\textwidth}  \textbf{ChatGPT}: "勾三股四弦五"是指中国传统乐器古琴的一种调式。  \end{varwidth} \\
          \cline{2-5}
       &   raising questions 
       &  \xmark & \cmark & ChatGPT usually \textbf{cannot} raise questions in response to patients 
       \\
        \hline
         \multirow{3}{*}{patient-friendly}&informativeness    &  \textbf{ high} & - & ChatGPT usually provides informative responses \\
        \cline{2-5}
        &patience  & \textbf{high} & - & ChatGPT has patience to provide lengthy  responses \\
        \cline{2-5}
       & presentation quality  &  \textbf{high} & -   & ChatGPT usually provide  well-formatted responses (like using bullets)  \\
        \bottomrule
    \end{tabular}
    }
    \caption{Behavior difference between ChatGPT and Doctors in various aspects. We argue that learning from ChatGPT and Doctors are complementary.  }
    \label{tab:comparision}
\end{table}


\paragraph{Doctors' Responses}
When consulting with doctors about our medical conditions, their responses typically exhibit professionalism that meets the personalized consultation. They are adept at inquiring about the symptoms and providing accurate diagnoses. However, due to time constraint~\footnote{Interestingly, ChatGPT does not has a sense of time and life, it does not need to save time.}, their replies are often informal and concise in nature, and sometimes incoherent. Our preliminary study shows that training from purely patient-doctor interaction data is not satirised: 1) it cannot fluently follow diverse instructions or chats ; 2) the responses are short, poorly-presented, and sometimes uninformative, which are not patient-friendly.

\paragraph{ChatGPT Responses}
On the other side, although ChatGPT usually generates informative, well-presented and logical responses, it usually tends to enumerate multiple possibilities and provides general and high-level advice. Since ChatGPT does not raise questions and guide patients to describe their symptoms, it lacks patients' input that can be used to generate specialized responses. 
In general, its responses often lack the contextual understanding that a doctor possesses, resulting in abstract responses that offer little substantial help to patients.
In Conclusion, \textbf{ChatGPT does not perform like doctors that conduct interactive diagnosis}.

\subsection{Our Solution}
Considering these challenges, we propose to combine the strengths of both distilled data (from ChatGPT) and real-world data (from Doctors), as illustrated in Table~\ref{tab:Comparison of Data Sources and Training Method Across Popular Medical Models}. The objective is to tame the medical LLM to perform like doctor. For example, it is expected to not only provide detailed, informative, and well-presented content but also conduct accurate and interactive diagnostic (usually posing clarifying questions) like doctors. To this end, our approach first mix distilled and real-world data in the Supervised Fine-Tuning stage (SFT). Furthermore, we employ RL from AI Feedback (RLAIF) to leverage the strengths of both data and meanwhile mitigate their weaknesses. 

\begin{table}
\centering
\resizebox{\textwidth}{!}{
\begin{tabular}{lccccccc}
\toprule
\multirow{2}{*}{\textbf{Model}}
&  \multirow{2}{*}{\textbf{Language}}  &
\multicolumn{2}{c}{\textbf{Instruction Data}}
&
\multicolumn{2}{c}{\textbf{Conversation Data}}
 &  \multirow{2}{*}{\textbf{Training Method}}  \\
 &  & Distilled & Real-world & Distilled & Real-world &  \\
\midrule
ChatDoctor & English & \cmark & \cmark & - & - & SFT \\
MEDALPACA & English & \cmark & \cmark & - & - & SFT \\
Visual Med-Alpaca & English & \cmark & \cmark & - & \cmark & SFT  \\
\hline
BenTsao & Chinese & \cmark & - & - & - & SFT \\
DoctorGLM& Chinese & \cmark & \cmark & - & \cmark & SFT \\
\hline
HuatuoGPT (Ours) & Chinese & \cmark & \cmark & \cmark & \cmark & SFT+RLAIF \\
\bottomrule
\end{tabular}}
\caption{Comparison of Data Sources and Training Method Across Popular Medical Models.}
\label{tab:Comparison of Data Sources and Training Method Across Popular Medical Models}
\end{table}

\section{Methodology} 

\begin{figure}[h]
  \centering
  \includegraphics[width=0.8\columnwidth]{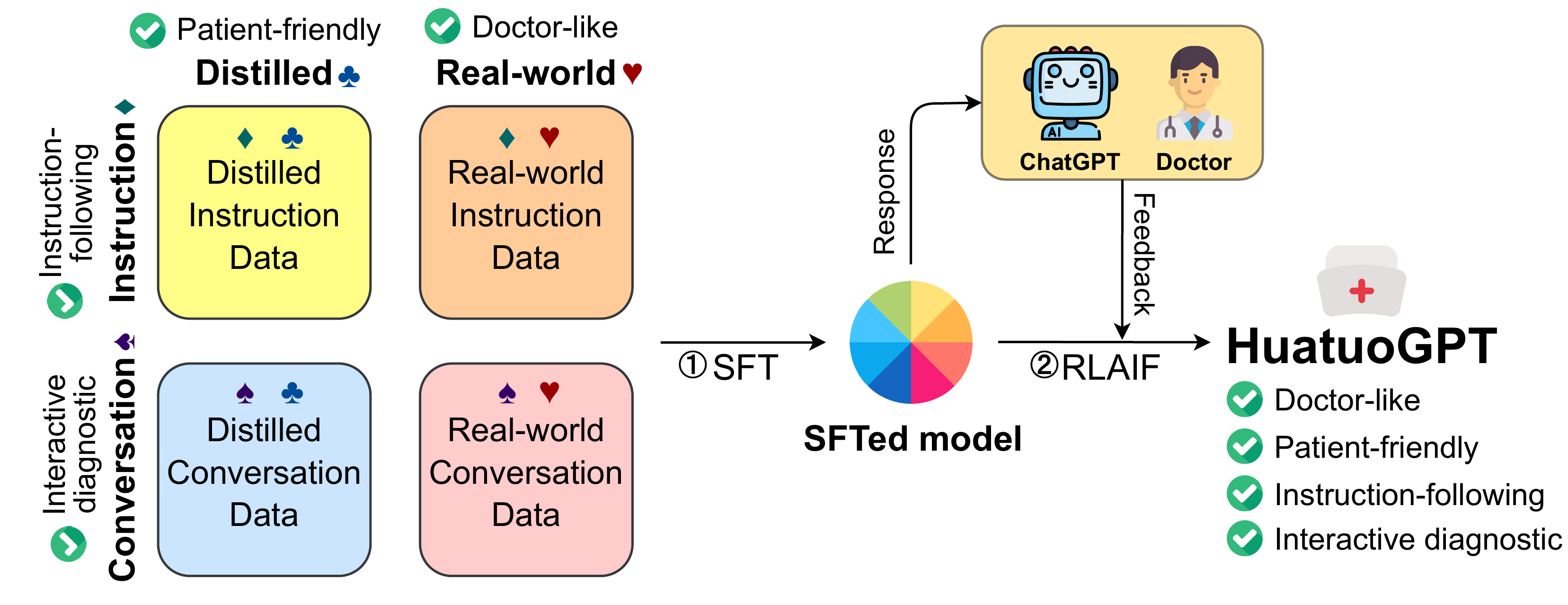}
  \caption{\label{fig:method}Schematic of HuatuoGPT.  }
\end{figure}

Our approach focuses on integrating the characteristics of both doctor and ChatGPT to enhance the quality of responses in medical consultations through a two-stage training strategy: SFT with hybrid data and RL with AI feedback. We first utilize well-selected hybrid data to train the model through supervised fine-tuning and subsequently reinforce the generation of desired responses through feedback from AI, as illustrated in Figure ~\ref{fig:method}.


\subsection{SFT with Hybrid Data}

In the first stage, we employ a blend of distilled data and real-world data, capitalizing on both strengths to endow the model with Doctor-like and Patient-friendly characteristics. Within each data category, we have collected instruction data and conversation data to imbue the model with the capacity for instruction-following and interactive diagnosis.



\begin{table}
\centering
\resizebox{\textwidth}{!}{
\begin{tabular}{lllp{3cm}p{8cm}}
\toprule
Data      &   Sources              & \multicolumn{1}{c}{Scale} & \multicolumn{1}{c}{Description}                                                                                                                                                                                    & \multicolumn{1}{c}{Example}                                                                                                                                                                                                                                                                                                                                                                                  \\ 
\midrule
\multirow{17}{*}{Instruction} &  \multirow{8}{*}{ChatGPT}    &     \multirow{8}{*}{61,400 }               & Medical instructions generated with self-instruction. Seed instructions are manually written.                                                                                                                      & User:现在该做什么准备手术前的身体检查？\newline Doctor:在准备手术前的身体检查中，需要进行以下几个步骤：\newline 1. 了解病人的病史：了解病人的病史，包括既往病史、手术史、药物过敏史、家族病史等，以便评估手术风险和制定手术方案。\newline 2. 进行体格检查：进行全面的体格检查，包括测量身高、体重、血压、心率、呼吸… \\
\cline{2-5}
&  \multirow{9}{*}{Doctors}   &    \multirow{9}{*}{69,768 }                  & Single-turn medical consultants in real-world scenarios, in which the doctor's response is more specific and can address the user's medical concerns.                                                              & User: 我有糖尿病血压160-90.该吃什么降压药？空腹血糖8左右。餐后11左右。我有糖尿病血压160-90.该吃什么降压药？\newline Doctor: 阅读你的提问描述，综合分析症状，体征，糖尿病，血压160-90.血压略高，空腹血糖8左右。餐后11左右。血糖略高。口服依那普利，二甲双胍片治疗，新近研究，二甲双胍片，可以抑制食欲，促进肠道蠕动，预防癌症，预防心脑血管病，延长寿命作用。                                                                                                                                                                  \\
\hline
\multirow{28}{*}{Conversation} &   \multirow{14}{*}{ChatGPT}   &    \multirow{14}{*}{ 68,888  }                & Medical multi-turn conversations are created using two ChatGPTs: one acting as the patient and the other as the doctor. The conversations are based on real-world patient cases and their corresponding treatment. & User: 我想问一下大夫，我发烧三天后，第四天全身起了疹子，是怎么回事呢？\newline Doctor:~您好！感谢您的提问。您的症状可能与幼儿急疹相关。幼儿急疹是一种常见的儿童病，病因为人类疱疹病毒6型或7型感染引起。它的症状包括：\newline 1. 发热 n2. 全身起红色疹子 \newline 3. 口腔内及咽喉部有小水泡\newline4. 病程一般为3-5天左右...\newline User:~谢谢大夫！除了幼儿急疹还有其他可能吗？我之前有感冒，这和疹子有关系吗？\newline Doctor:~除了幼儿急疹，病毒性感冒和其他病毒感染也有可能引起疹子，但这种情况相对较少见。在您的情况下，感冒可能是一种诱发因素，但主要还是由病毒感染引起的疹子...  \\
\cline{2-5}
&  \multirow{16}{*}{Doctors} &     \multirow{16}{*}{25,986 }                  & Medical multi-turn conversations in real-world scenarios. Conversations include multiple rounds of questioning by doctors and continuous questioning by patients.                                                  & User：额头下巴下长满痘痘。就是不知道.\newline Doctor：你好，有什么困扰吗？我可以为你提供帮助。请问您平时的饮食习惯、睡眠和护肤习惯怎样？化妆品和防晒霜的使用频率如何？还有洁面的频次是怎样的呢？\newline User：平时我基本上只是用水乳霜，然后洗面奶只用一种。日常防晒只有在夏天时会使用，我一般用悦诗风吟的洗面奶以及AHC的护肤产品，这些都比较温和。 \newline Doctor：谢谢您的描述。首先，建议您停用化妆品和普通护肤品，使用特定的药妆品牌，如雅漾和理肤泉的洁面泡沫或洁面乳，并避免使用热水或肥皂清洗面部。建议您外用阿达帕林凝胶涂于面部痤疮，一般夜间使用，涂抹十五分钟之后外用医学护肤品比如雅漾、薇资、理肤泉清爽型的舒缓面霜，或者是维生素e乳膏….                               \\
\bottomrule
\end{tabular}
}
\caption{The Details of Hybrid SFT Data.}
\label{tab:Hybrid Data}
\end{table}

\paragraph{Distilled Instructions from  ChatGPT}
We follow the work of self-instruct~\cite{wang2022self,alpaca,chen2023phoenix} to construct a set of medical instruction data aiming to enable the model to follow user's medical instructions. The difference is that we have employed top-down manner to create more natural and comprehensive responses. We design a taxonomy to collect or manually create seed instructions based on the roles and use cases. Based on each role or use case, we generate instructions separately using self-instruct~\cite{wang2022self}. This could provide a wide range of instructions and meanwhile keep enough instructions for each role or use cases. Finally, we mix all seed instructions together and conduct self-instruct; this might be helpful to generate more diverse instructions. Details refer to Appendix~\ref{Prompt_single_distilled}.


\paragraph{Real-world Instructions from Doctors}
Real-world instruction data are derived from  question-answering between doctors and patients. Responses from doctors are expertise, with high relevance and conciseness. Therefore, we further enhance the quality and reliability of the single-turn instruction data by refining authentic doctor-patient question-answer pairs. Details refer to Appendix~\ref{Prompt_single_real}.

\paragraph{Distilled Conversations from ChatGPT}
Distilled conversations are generated by  two ChatGPTs, each ChatGPT is associated with a role (either doctor or patient) using a well-designed prompt. First, we leverage a third-party medical diagnosis database as a valuable source of medical knowledge and expertise for generating synthetic dialogue data. Based on the basic background of patients and the final  diagnosis from  doctors,  two ChatGPTs are asked to generate dialogue utterances one by one. 
In these conversations, the responses generated by LLMs usually are informative, detailed, well-presented,  and adhere to a consistent style; the format and information  are usually friendly to patients.  Details refer to Appendix~\ref{Prompt_conv_real}


\paragraph{Real-world Conversations with Doctors}
Real-world conversations are collected from genuine scenarios, where doctors' responses often demand diverse abilities, including long-range reasoning and raising questions to guide patients in describing their symptoms.
However, this type of data sometimes suffers from being overly concise and too colloquial. To address this, we utilized language models to enhance and refine the data based on the original content, which yields a high-quality real conversation dataset. Details refer to Appendix~\ref{Prompt_conv_real}

\subsection{RL with AI Feedback}
In the Supervised Fine-Tuning (SFT) phase, we introduced a diverse dataset with the aim of enabling HuatuoGPT to emulate 
the inquiry and diagnosing strategy of doctors,
while maintaining the rich, logical, coherent characteristics of LLMs' responses. In order to further align the model's generation preferences to our needs, we propose reinforcement learning with AI feedback to improve the quality of models' responses. Previously, OpenAI introduced reinforcement learning with human feedback\cite{ouyang2022training} to align LLMs with human preference but at a significant time and labor cost. \cite{bai2022constitutional} demonstrated that with a carefully designed prompt, AI is able to imitate human preferences and to give relatively consistent scores on generated responses. Inspired by these alignment methods, we design a new pipeline to force the model to generate informative and logical responses without deviating from doctor's diagnosis.

\paragraph{Reward Model}
We train a reward model to align with the characteristics of doctors and LLMs. We use real instructions and conversations as training data, sampling multiple responses from our fine-tuned model. For multi-turn conversations, we provide the dialogue history to align our model's response generation. These responses are then scored by an LLM, such as ChatGPT, considering informativeness, coherence, adherence to human preferences, and factual accuracy based on given real doctors' diagnoses. The scoring LLM evaluates each response and assigns a score. We use this paired response data to train the reward model, using the fine-tuned model as its backbone for better generalization. 

\paragraph{Reinforcement Learning}
In RL process, we sample $k$ different responses 
$\{y_1, \dots, y_k\}$
of a given query $x$ by current policy $\pi$. Each response $y_i$ is fed to our reward model to provide a reward score $r_{RM}$. To ensure that the model does not deviate too far from the initial state $\pi_0$, we add the empirically-estimated KL penalty term, and the final reward function is as follows:
\begin{equation}
    r = r_{RM} - \lambda_{KL}D_{KL}(\pi||\pi_0)
\end{equation}
where $\lambda_{KL}$ is a hyperparameter for KL penalty, $D_{KL}$ is the KL penalty function. In our experiment, $\lambda_{KL}$ is set to $0.05$. 
Input queries are de-duplicated and sampled from the remaining SFT hybrid data. This ensures a diverse range of inputs while retaining the model's response preferences in both the single-turn instruction and the multi-turn conversation scenarios.


\section{Experiments}
In this section, we first introduce the training implementation (Section~\ref{sec:training_details}) and then present the evaluation manners and results including automatic evaluation (Section~\ref{sec:auto_eval}) and manual evaluation (Section~\ref{sec:manual_eval}). 

\subsection{Training Details}\label{sec:training_details}
Our model is implemented in PyTorch using the Accelerate\footnote{\url{https://huggingface.co/docs/accelerate/index}} and trlx\footnote{\url{https://github.com/CarperAI/trlx}} packages with Bloomz-7b1-mt~\cite{muennighoff2022crosslingual} as the base architecture.\footnote{We adopt BLOOMZ-7b1-mt as our backbone, which stands out with its exceptional multilingual capabilities and suitable for open-source applications. BLOOMZ model family is trained with the PILE corpus \cite{gao2020pile}, which contains varied medical texts, including resources like PubMed Central and PubMed Abstracts. These valuable texts significantly enrich the BLOOMZ models with an extensive body of medical knowledge, subsequently enables our models to perform better in the medical domain.} We leverage ZeRO-3~\cite{rajbhandari2020zero} to distribute the model across 8 A100 GPUs for training. In the supervised fine-tuning process, we set the learning rate, batch size, and maximum context length to $2e-5$, $128$, and $2048$, respectively. All models are trained for 3 epochs and weights performed the best on the validation set are saved. During the reinforcement learning process, we only update the parameters of the last two layers. The total number of steps is $16,000$, with a learning rate of $8e-6$.
%
%
In addition, to enhance the model's conversational and instruction-following capabilities in the general domain, we have incorporated Chinese instruction data (the Chinese Alpaca dataset~\cite{peng2023instruction} and conversation data (ShareGPT\footnote{\url{https://huggingface.co/datasets/philschmid/sharegpt-raw}}). This enhances the model's ability to effectively understand and generate responses in various conversational scenarios and accurately follow instructions across different domains.

\subsection{Automatic Evaluation}\label{sec:auto_eval}

\subsubsection{Medical Benchmarks}

We select three existing Chinese medical QA datasets as examples, namely cMedQA2~\cite{8548603}, webMedQA~\cite{he2019applying} and Huatuo-26M~\cite{li2023huatuo}, and compare the results with the existing baselines. 
\textbf{cMedQA2} is a publicly available dataset based on Chinese medical questions and answers consisting of 108,000 questions and 203,569 answers. \textbf{webMedQA} is a real-world Chinese medical QA dataset collected from online health consultancy websites consisting of 63,284 questions. \textbf{Huatuo-26M}~\cite{li2023huatuo} is the largest Chinese medical QA dataset which has 26M QA pairs from online medical consultation, knowledge bases and encyclopedias.


\paragraph{Evaluation Metrics}
Following the previous works~\cite{li2023huatuo}, we utilize evaluation metrics such as BLEU, ROUGE, GLEU, and Distinct. BLEU computes the k-gram overlap between generated and reference sentences to measure similarity. ROUGE-N assesses the N-gram overlap, and ROUGE-L gauges the longest common subsequence of word matches. GLEU auto-evaluates sentence-level fluency. Distinct-1/2 aids in assessing textual diversity of the generated response by determining distinct n-grams count. However, these reference-based metrics may not suit medical QA scenarios due to diverse potential reference answers; more sound metrics should be paid more attention.

\paragraph{Baselines}
We compare our model to the best reported zero-shot model \textbf{ChatGPT} (GPT-3.5-turbo) and an in-domain fine-tuned model \textbf{Chinese T5}~\footnote{\url{https://huggingface.co/imxly/t5-pegasus}} respectively, which is continuously trained for 1 epoch on the full training set using batch-size 8, with a learning rate of $10^{-4}$ using Adam, linear scheduling with a warm-up rate of 0.1. 


\begin{table*}[t]\footnotesize
\centering
\addtolength\tabcolsep{-4.5pt} 
 \resizebox{\textwidth}{!}{
\begin{tabular}{llcccccccccccc}
\toprule
Dataset & Model & BLEU-1 & BLEU-2 & BLEU-3 & BLEU-4 & GLEU  & ROUGE-1 & ROUGE-2 & ROUGE-L & Distinct-1 & Distinct-2 \\ 
\midrule

\multirow{4}{*}{\textbf{cMedQA2}}&GPT-3.5-turbo & 19.21 & 7.43 & 3.14 & 1.24 & 5.06 & 20.13 & 3.10 & 12.57 & 0.69& \textbf{0.99} \\
&T5 (fine-tuned) $^ \dagger$ & 20.88   & 11.87   & 7.69    & \textbf{5.09}    & 7.62  & 27.16    & \textbf{9.30}     & \textbf{20.11}    & 0.41& 0.52\\
& \textbf{HuatuoGPT} & \textbf{25.37} & \textbf{13.16} & \textbf{7.39} & 4.25 & \textbf{8.30} & \textbf{27.75} & 7.31 & 17.36 & \textbf{0.74} & 0.93 \\
\midrule

\multirow{4}{*}{\textbf{webMedQA}}  &GPT-3.5-turbo & 18.06 & 6.74 & 2.73 & 1.09 & 4.71 & 20.01 & 2.81 & 12.58 & 0.65 & 0.87\\
&T5 (fine-tuned)  $^ \dagger$  &  21.42   & 13.79   & \textbf{10.06}   & \textbf{7.38}    & \textbf{8.94}  & \textbf{31.00 }   & \textbf{13.85}    & \textbf{25.78}    & 0.37& 0.46\\
& \textbf{HuatuoGPT}& \textbf{24.61} & \textbf{12.84} & 7.23 & 4.19 & 7.73 & 27.38 & 7.09 & 17.66 & \textbf{0.71}& \textbf{0.93} \\
\midrule

\multirow{4}{*}{\textbf{Huatuo-26M}} 
&GPT-3.5-turbo & 18.44 & 6.95 & 2.87 & 1.13 & 4.87 & 19.60 & 2.82 & 12.46 & 0.69 & 0.89 & \\
&T5 (fine-tuned)&\textbf{26.63} &\textbf{16.74} &\textbf{11.77} &\textbf{8.46} &	\textbf{11.38} &\textbf{33.21} &\textbf{13.26} &\textbf{24.85} &0.51 &0.68 \\
&\textbf{HuatuoGPT} & 25.16 & 13.21 & 7.54 & 4.40 & 8.37 & 27.76 & 7.45 & 17.99 & \textbf{0.73} & \textbf{0.93} & \\
 \bottomrule
\end{tabular}}
 \caption{Benchmark on Chinese medical QA dataset~\cite{li2023huatuo}. GPT-3.5-turbo (ChatGPT) and HuatuoGPT are \textbf{zero-shot} while T5 is finetuned.  } 
\label{tab:generation_results} 
\end{table*}

\paragraph{Results} HuatuoGPT demonstrates impressive performance across various Chinese medical benchmarks, achieves consistently high scores across all metrics, and demonstrates a high level of accuracy, fluency, and diversity in its generated responses. In cMedQA2 and webMedQA, HuatuoGPT even outperforms fine-tuned T5, suggesting that it has a robust generalization capability and is able to effectively handle a wide range of medical question-answering tasks.





\subsubsection{Evaluation with GPT4}

We conduct an automated evaluation on single-turn questions with different intents and multi-turn conversations from different departments to observe the performance of the model in various scenarios. 

\paragraph{Evaluation dataset} For the single-turn questions, we extract 100 questions representing 10 intents (condition diagnosis, etiological analysis, treatment plan, medical advice, indicators interpretation, disease description, consequences description, precautions, efficacy, medical expenses) from the validation set of the Knowledge-based Universal Automated Knowledge Extraction for Query Intent Classification (KUAKE-QIC) in Chinese Biomedical Language Understanding Evaluation (CBLUE~\cite{zhang2021cblue})\footnote{\url{https://github.com/CBLUEbenchmark/CBLUE}}. KUAKE-QIC is collected from search engine queries, which makes it suitable for single-turn questions. To filter the noisy data, these questions were initially scored by ChatGPT, and a manual filtering process was conducted to select higher quality candidate questions for the test set. For the multi-turn questions, we used the patient cases from \cite{zeng2020meddialog}. We selected 20 departments and randomly sampled 5 patient cases from each department, resulting in a total of 100 real patient cases. These cases were provided to ChatGPT, which played the role of the patient, interacting with each doctor model to obtain the diagnosis results.


\paragraph{Evaluation aspects and metrics} We use GPT-4 as the referees to review the quality of model outputs. We prompt it to consider doctor-like language, symptom inquiry capability, the effect and reliability of the treatment recommendations and prescriptions, and the helpfulness to the patient. Given the question and the corresponding two answers from two models, GPT-4 is asked to first compare the advantages of each output and analyze the helpfulness to the patient, then it is requested to provide a score to each response respectively. In this way, we can get the evaluation scores of the 100 questions for each model comparison pair. We take the average scores over all the questions and calculate the performance ratio for each compared model (i.e. the overall score of the compared model divided by that of HuatuoGPT in a comparison pair). 


\paragraph{Baselines} We mainly compare HuatuoGPT to the two most popular general models ChatGPT and GPT4~\footnote{ChatGPT and GPT-4 version is the online one on 12th May 2023}, and the two most representative open-source Chinese medical large language models: BenTsao (tuned from LLaMA)\footnote{\url{https://github.com/SCIR-HI/Huatuo-Llama-Med-Chinese/tree/main}}, DoctorGLM (tuned from ChatGLM)\footnote{\url{https://github.com/xionghonglin/DoctorGLM}}. For single-turn questions evaluation, we compare to all the mentioned four models. For multi-turn conversations evaluation, we only compare our model to DoctorGLM and GPT-3.5-turbo due to the quote limit of GPT-4. We report the performance ratio of all models over all single-turn questions and multi-turn conversations respectively. 


\begin{table}[ht]
\centering
\resizebox{\textwidth}{!}{
\begin{tabular}{l|cc|cc|cc|cc}
\toprule
\textbf{\small Category} & \multicolumn{2}{c|}{\textbf{\scriptsize HuatuoGPT v.s. BenTsao}} & \multicolumn{2}{c|}{\textbf{\scriptsize HuatuoGPT v.s. DoctorGLM}} & \multicolumn{2}{c|}{\textbf{\scriptsize HuatuoGPT v.s. ChatGPT}} & \multicolumn{2}{c}{\textbf{\scriptsize HuatuoGPT v.s. GPT-4}} \\
\midrule
{\small Efficacy} & 8.60 & 2.40 & 8.30 & 4.15 & 7.50 & 7.45 & 7.00 & 7.05 \\ 
{\small Medical Expenses} & 8.50 & 3.30 & 8.20 & 4.10 & 8.05 & 6.65 & 7.20 & 7.45 \\ 
{\small Consequences Description} & 8.70 & 4.00 & 8.60 & 3.60 & 7.50 & 7.80 & 7.20 & 8.65 \\ 
{\small Medical Advice} & 8.60 & 3.90 & 8.70 & 3.50 & 7.25 & 8.05 & 6.65 & 9.00 \\ 
{\small Indicators Interpretation} & 8.60 & 3.40 & 8.70 & 4.50 & 8.55 & 6.50 & 7.40 & 8.30 \\ 
{\small Treatment Plan} & 8.90 & 3.40 & 8.40 & 4.00 & 7.45 & 8.30 & 6.80 & 8.95 \\ 
{\small Precautions} & 9.00 & 4.20 & 9.00 & 4.80 & 7.70 & 8.25 & 6.60 & 8.45 \\ 
{\small Disease Description} & 8.60 & 2.70 & 8.75 & 3.80 & 7.75 & 7.65 & 7.45 & 8.15 \\ 
{\small Etiological Analysis} & 8.70 & 3.40 & 8.50 & 3.40 & 7.65 & 8.20 & 7.00 & 8.50 \\ 
{\small Condition Diagnosis} & 8.60 & 4.05 & 8.40 & 5.05 & 7.90 & 7.55 & 7.30 & 8.10 \\ 
\hline
{\small Overall} & \textbf{8.68} & 3.47 & \textbf{8.56} & 4.09 & \textbf{7.73} & 7.64 & 7.06 & \textbf{8.26} \\ 
\bottomrule
\end{tabular}
}
\caption{\label{tab:gpt4_review}Scores for each category and the overall scores of each model pair on 100 single-turn questions reviewed by GPT-4.}
\end{table}

\begin{table}[ht]
\centering\small
\begin{tabular}{l|cc|cc}
\toprule
\textbf{Category}                                         & \multicolumn{2}{c|}{\textbf{\small HuatuoGPT v.s. DoctorGLM}} & \multicolumn{2}{c}{\small \textbf{HuatuoGPT v.s. ChatGPT}} \\
\midrule
Traditional Chinese Medicine               & 8.80                      & 5.70                      & 8.60                     & 8.30                     \\
Obstetrics                                 & 9.00                      & 5.20                      & 8.90                     & 7.50                     \\
Pediatrics                & 8.90                      & 6.50                      & 8.50                     & 7.80                     \\
Internal Medicine                                                    & 8.80                      & 5.60                      & 8.70                     & 7.70                     \\
Stomatology               & 8.60                      & 6.60                      & 8.30                     & 8.40                     \\
Surgery                      & 8.00                      & 5.20                      & 8.60                     & 7.60                     \\
Obstetrics and Gynecology & 8.70                      & 5.80                      & 8.60                     & 7.60                     \\
Gynecology                                                    & 8.50                      & 6.40                      & 8.10                     & 8.00                     \\
Cardiovascular Medicine   & 8.60                      & 5.20                      & 8.40                     & 8.00                     \\
General Surgery                                                    & 9.00                      & 5.00                      & 8.80                     & 7.70                     \\
Urology                   & 8.70                      & 5.80                      & 8.50                     & 7.40                     \\
Gastroenterology                                                    & 8.70                      & 5.50                      & 8.70                     & 7.60                     \\
Andrology                 & 8.90                      & 5.20                      & 8.20                     & 7.50                     \\
Dermatology and Venereology                                                    & 8.70                      & 5.80                      & 8.40                     & 7.60                     \\
Dermatology               & 8.80                      & 5.30                      & 7.90                     & 7.70                     \\
Ophthalmology                                                    & 8.50                      & 6.00                      & 7.60                     & 8.40                     \\
Neurology                 & 8.80                      & 5.90                      & 9.00                     & 8.10                     \\
Neurosurgery                                                    & 8.80                      & 4.20                      & 8.10                     & 8.40                     \\
Psychiatry                & 8.70                      & 5.80                      & 8.20                     & 7.70                     \\
Orthopedics                                                    & 8.80                      & 5.90                      & 8.30                     & 8.00                     \\
\hline
Overall                                    & \textbf{8.72}                      & 5.63                      & \textbf{8.42}                     & 7.85             \\       
\bottomrule
\end{tabular}
\caption{\label{tab:gpt4_review_dialogue}Scores for each category and the overall scores of each model pair on 100 multi-turn dialogues reviewed by GPT-4.}
\end{table}

\begin{figure}[h]
  \centering
  \includegraphics[width=0.8\columnwidth]{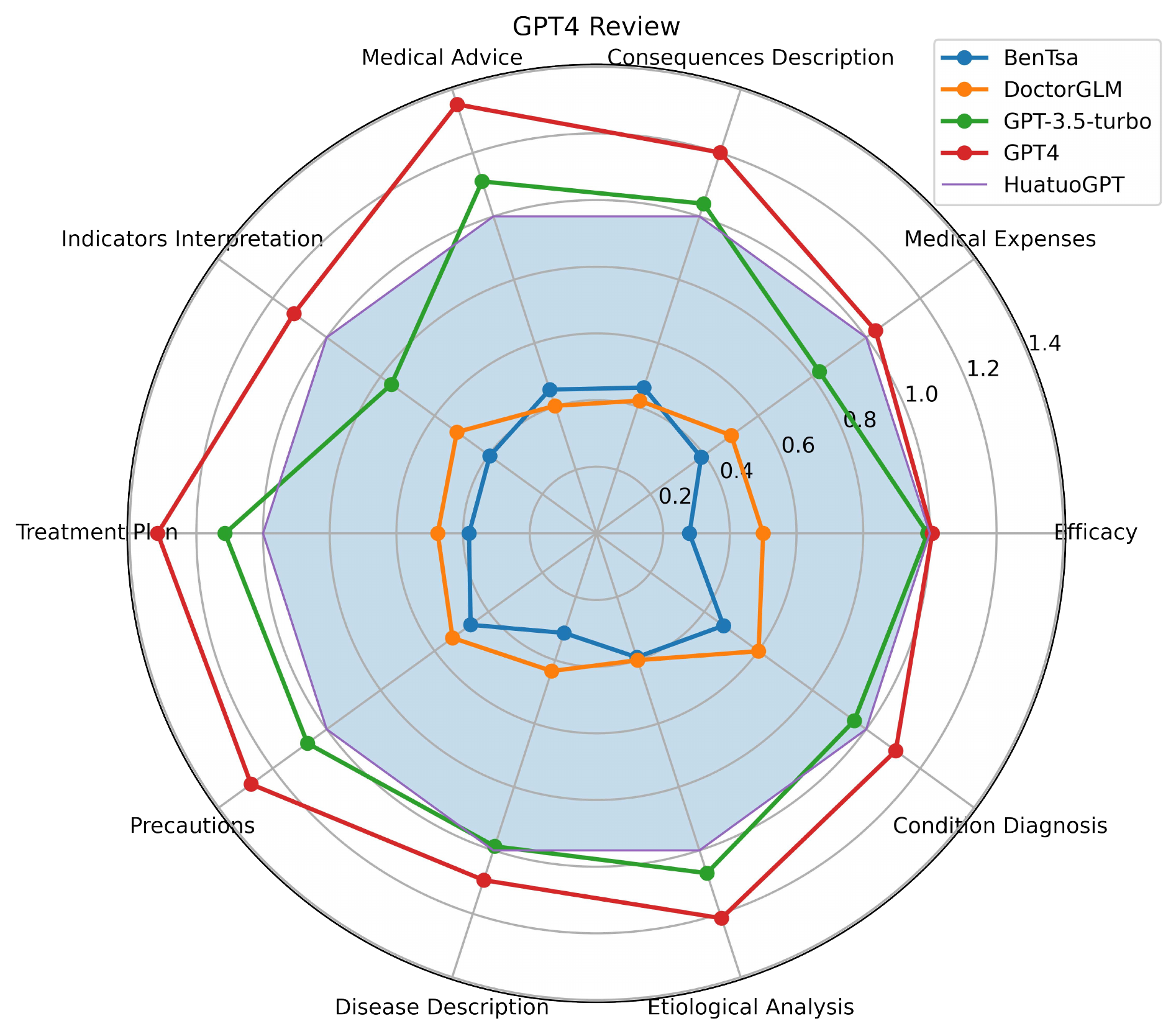}
  \caption{\label{fig:gpt4_review}The model performance ratio for each category on 100 single-turn questions, reviewed by GPT4. The performance of HuatuoGPT is set to 1.0.}
\end{figure}
\begin{figure}[h]
  \centering
  \includegraphics[width=0.7\columnwidth]{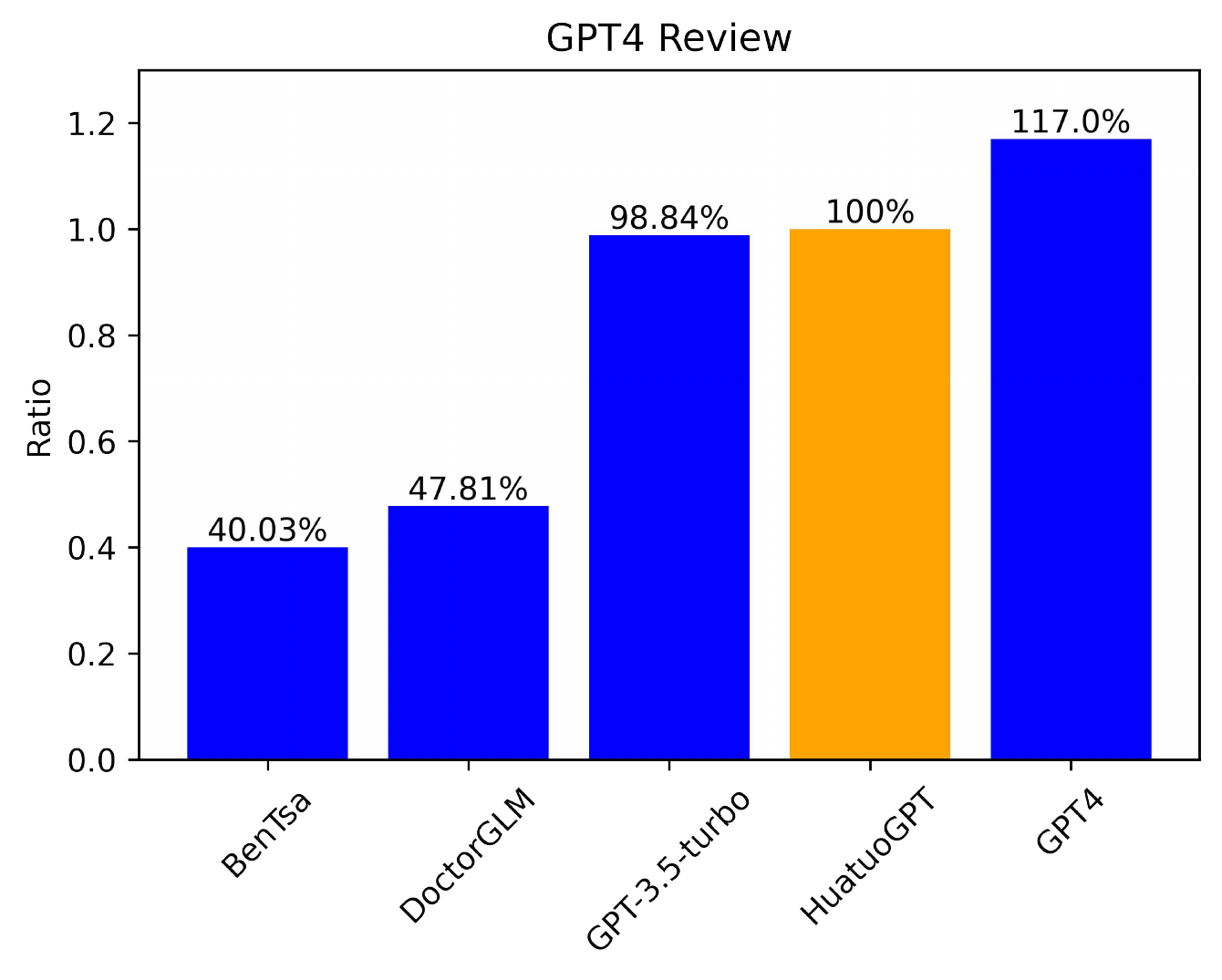}
  \caption{\label{fig:gpt4_review_total}The overall model performance ratio on 100 single-turn questions, reviewed by GPT-4. The performance of HuatuoGPT is set to 100\%.}
\end{figure}
\begin{figure}[h]
  \centering
  \includegraphics[width=0.8\columnwidth]{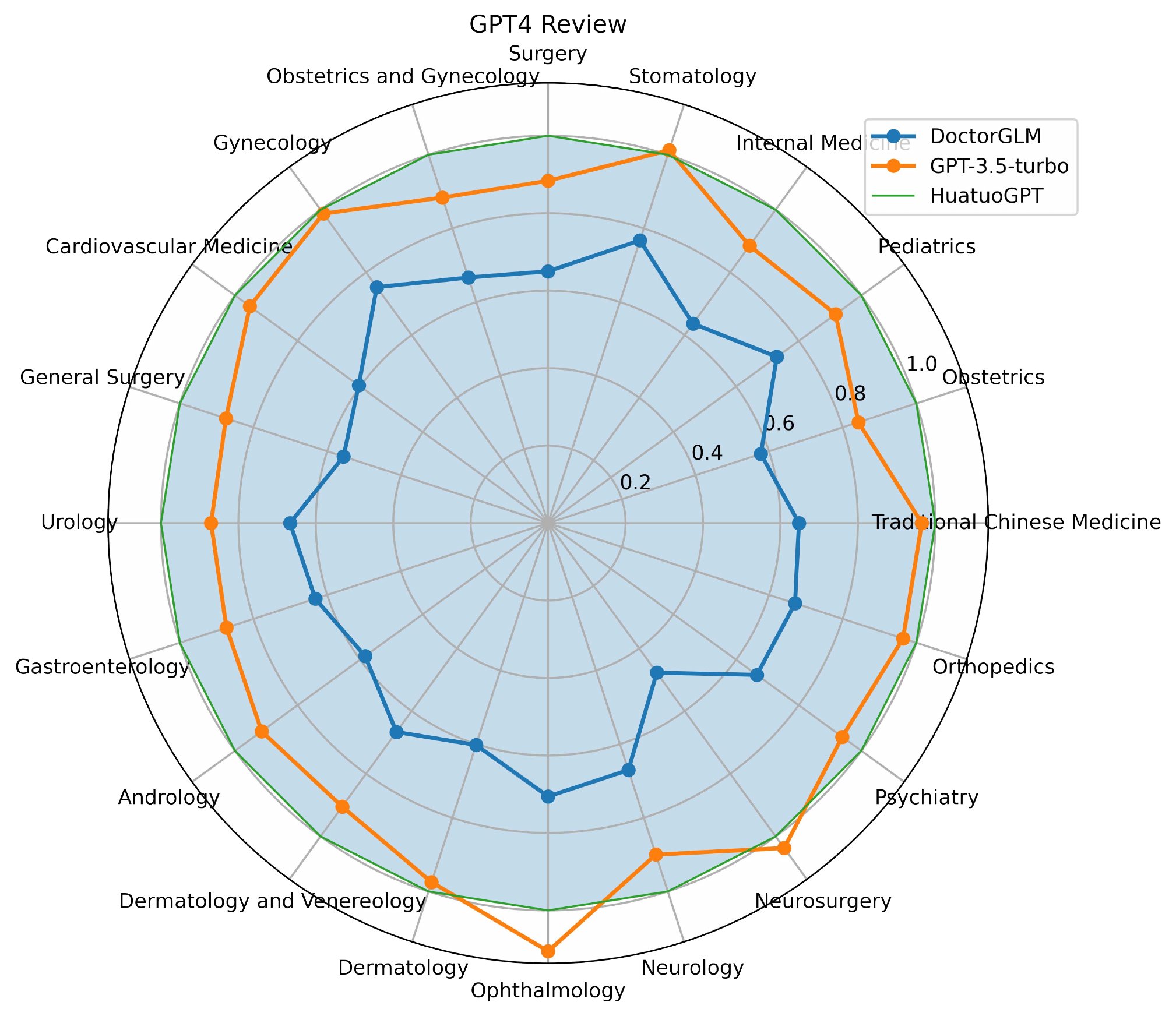}
  \caption{\label{fig:gpt4_review_dialogue}The model performance ratio for each category on 100 multi-turn dialogues, reviewed by GPT4. The performance of HuatuoGPT is set to 1.0.}
\end{figure}
\begin{figure}[h]
  \centering
  \includegraphics[width=0.7\columnwidth]{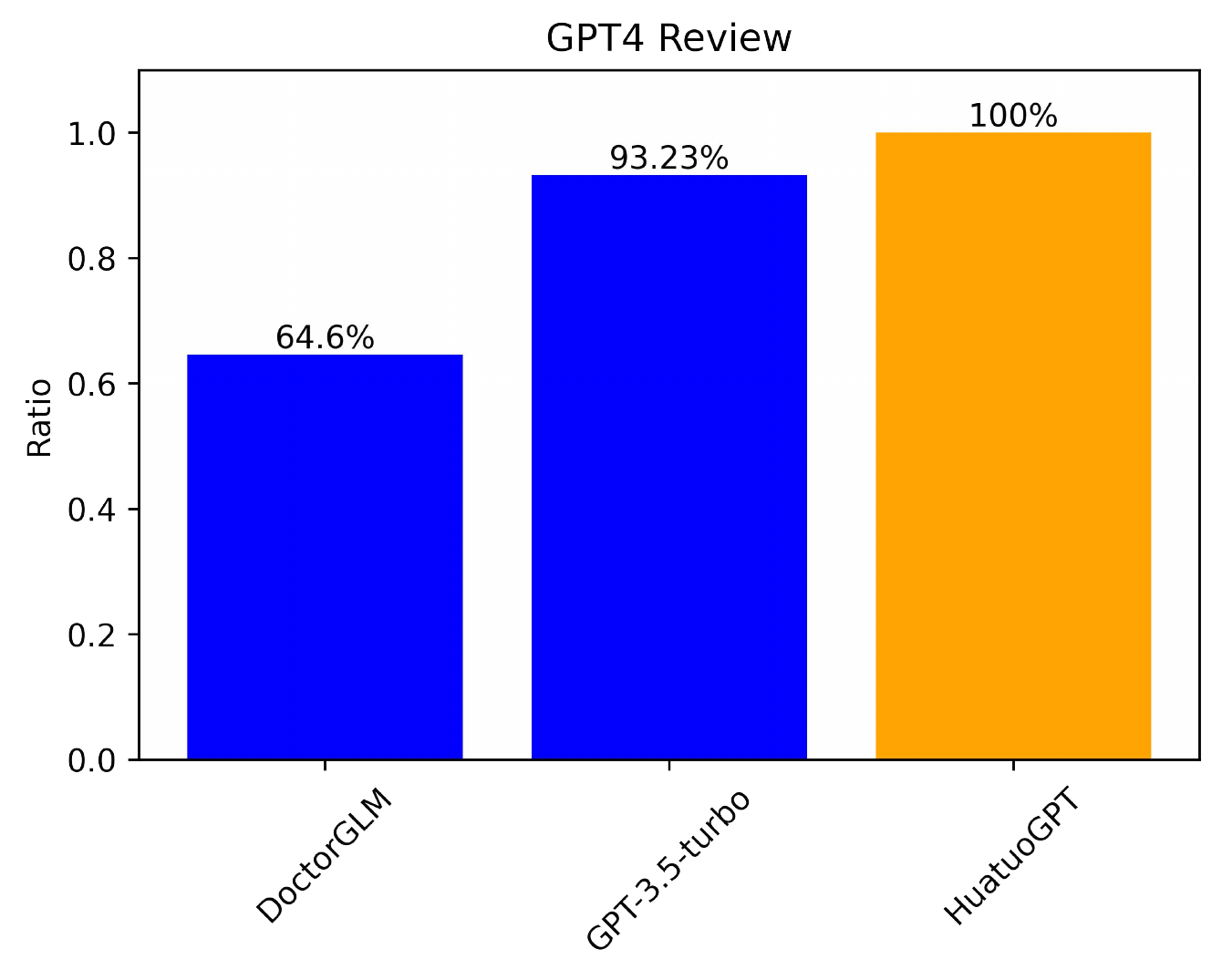}
  \caption{\label{fig:gpt4_review_total_dialogue}The overall model performance ratio on 100 multi-turn dialogues, reviewed by GPT-4. The performance of HuatuoGPT is set to 100\%.}
\end{figure}

\paragraph{Results} For the single-turn questions evaluation, all the model performance results are shown in Table~\ref{tab:gpt4_review}. The comparison among models for each category is shown in Figure~\ref{fig:gpt4_review} and the comparison among their overall performance is shown in Figure~\ref{fig:gpt4_review_total}, where the performance of HuatuoGPT is set to 1.0. According to GPT-4, HuatuoGPT is much better than BenTsao and DoctorGPT in all categories. Compared to GPT-3.5-turbo, HuatuoGPT outperforms it in three categories (Indicators Interpretation, Condition Diagnosis, and Medical Expenses) and performs similarly to it in two categories (Efficacy and Disease Description). However, HuatuoGPT is still worse than GPT4 in almost all categories, where it attains similar performance to GPT4 in two categories (Efficacy and Medical Expenses). Overall, HuatuoGPT achieves higher scores than DoctorGLM, BenTsao, and GPT-3.5-turbo. For the multi-turn conversations evaluation, similarly, the overall performance of HuatuoGPT surpasses GPT-3.5-turbo in over 60\% of cases. The comparison for each category and for overall performance are shown in Table~\ref{tab:gpt4_review_dialogue}, Figure~\ref{fig:gpt4_review_dialogue}, and Figure~\ref{fig:gpt4_review_total_dialogue} respectively.


\subsection{Manual Evaluation}\label{sec:manual_eval}

\subsubsection{Evaluation dataset and metrics}
We utilize the 100 KUAKE-QIC questions (the same as those in automated evaluation) as the test set for single-turn question evaluation and randomly sample 50 patient cases from 100 test cases used in automated evaluation for multi-turn conversations manually evaluation.

In the manual evaluation of the HuatuoGPT, we think that the following three aspects should be considered, particularly in medical consultation and medication prescription and take them as the guidelines for evaluation:

\textbf{Diagnosis accuracy}. This aspect evaluates the model's accuracy and comprehensiveness in diagnosing patient symptoms. Evaluators are provided a set of medical cases or symptom descriptions and assess the correctness, relevance, and reasonableness of the model's diagnosis. Comparisons can be made with assessments made by medical professionals to ensure the model's accuracy.

\textbf{Treatment recommendation accuracy}. This aspect assesses the accuracy and appropriateness of the model's treatment recommendations for patients. Evaluators are provided a set of medical cases or symptom descriptions and evaluate whether the model's treatment recommendations align with medical knowledge and real-world applications that are effective and reliable to the patient's main condition and problem.

\textbf{Medication knowledge and prescription accuracy}. This aspect evaluates the model's understanding of medications and the accuracy of its prescription recommendations. Evaluators are provided a set of medical cases or symptom descriptions and assess the accuracy and reliability of the medication recommendations based on medical knowledge and guidelines.

We provide physicians with above considerations, enabling them to align their evaluation guidelines. This allows for a meticulous comparison of the good and bad outputs of different models for the same scenario.

During evaluation, medical experts are asked to provide assessments on different responses. Each physician is solely responsible for evaluating the output of a single pair of models, ensuring that each response data is scrambled and anonymized with the utmost strictness. Consistent with automatic evaluation, we take BenTsao, DoctorGLM, ChatGPT and GPT4 as the baselines in single-turn question evaluation and select DoctorGLM, ChatGPT as the baselines in multi-turn conversation evaluation.



\subsubsection{Results} 

As shown in Table~\ref{tab:single_human_eval}, HuatuoGPT performs exceptionally well against BenTsao, DoctorGLM, and it even slightly outperforms ChatGPT, highlighting its robust diagnosis accuracy, treatment recommendations, and medication knowledge. The results of multi-turn evaluation are shown in Table~\ref{tab:multi_human_eval}, which reveals that HuatuoGPT excels in extended dialogue contexts, evidenced by an 86\% win rate against DoctorGLM and 58\% against ChatGPT. It indicates that HuatuoGPT has a more prominent interactive diagnostic capability in patient consultation scenarios. 

\begin{table}[htbp]
    \centering
    \begin{tabular}{lccc}
    \hline
         Comparison & Win & Lose \\
         \hline
          \textbf{HuatuoGPT} vs BenTsao & 100\% & 0\% \\ 
         \textbf{HuatuoGPT} vs DoctorGLM & 98\% &  2\% \\  
         \textbf{HuatuoGPT} vs ChatGPT & 52\% &  48\% \\  
         \textbf{HuatuoGPT} vs GPT-4 & 10.5\% &  89.5\% \\  
        
         \hline
    \end{tabular}
    \caption{Manual evaluation of model performance in single-turn medical consultations.}
    \label{tab:single_human_eval}
\end{table}

\begin{table}[htbp]
    \centering
    \begin{tabular}{lccc}
    \hline
         Comparison & Win & Lose \\
         \hline
          \textbf{HuatuoGPT} vs DoctorGLM & 86\% & 14\% \\ 
         \textbf{HuatuoGPT} vs ChatGPT & 58\% &  42\% \\  
        
         \hline
    \end{tabular}
    \caption{Manual evaluation of model performance in multi-turn medical consultations.}
    \label{tab:multi_human_eval}
\end{table}


\section{Discussion}

\subsection{Ablation Study}
In this section, we explore the impact of two types of data on the model. We trained two distinct models, namely \textbf{HuatuoGPT (w/ real data)} and \textbf{HuatuoGPT (w/ distilled data)}, using exclusively real-world data or distilled data, respectively. We thoroughly compare the variations in responses between the two models for the same set of questions as shown in Table~\ref{case study}.
\textbf{HuatuoGPT (w/ real data)} has a tendency to ask clarifying questions to patients, performing as expected, similar to a doctor. However, a minor flaw is that the response is brief and the content appears less well-organized for reading.
On the other hand, \textbf{HuatuoGPT (w/ distilled data)} generates well-organized, detailed, and informative content. Nevertheless, its responses are more focused on providing suggestions rather than making a diagnostic decision. Thus, \textbf{HuatuoGPT (w/ distilled data)} resembles a "non-doctor friend" rather than a doctor.


To assess the impact of RLAIF (Reinforced Learning with Auxiliary Information Feedback), we also compare two models: the default model called \textbf{HuatuoGPT} and a variant called \textbf{HuatuoGPT (w/o RLAIF)} which does not utilize RLAIF. It is worth noting that the latter model, \textbf{HuatuoGPT (w/o RLAIF)}, did not ask additional questions to patients. This might be attributed to the fact that its training data could be biased towards the ChatGPT data, while real-world data may have been overlooked. In contrast, our default model, \textbf{HuatuoGPT with RLAIF}, can function like a doctor by asking follow-up questions to patients  to get more accurate diagnoses.

\begin{table*}[t]
\tiny
\centering
\resizebox{\textwidth}{!}{
\begin{tabular}{p{12cm}}
\toprule
\textbf{User:}\newline 我最近手上，手心长小泡，十个手指肚脱皮。想问这是什么病，需要去医院吗？\newline (I recently developed small blisters on my hands and palms, and my ten fingers are peeling. What is this disease and do I need to go to the hospital?)\\
\midrule
\textbf{HuatuoGPT (w/t real data):} \newline 您好，根据您的症状描述，您可能患有湿疹或接触性皮炎等皮肤疾病引起的过敏反应。为了更准确地诊断病情，我需要了解一些额外的信息。请问这些小泡是否伴有瘙痒感？是否有其他部位的皮肤出现类似的症状呢？\newline (Hello, based on the description of your symptoms, you may be suffering from an allergic reaction caused by a skin condition such as eczema or contact dermatitis. In order to diagnose the condition more accurately, I need to know some additional information. Are these small blisters accompanied by itching? Are there any other areas of the skin that show similar symptoms?)\\
\midrule
\textbf{HuatuoGPT (w/t distilled data):}\newline 您好，根据您描述的症状，可能是湿疹引起的皮肤炎症反应。但是为了更精确地确定病因和治疗方案，建议您到医院就诊。医生可以通过仔细检查您的症状和询问相关历史信息来做出明确的诊断。\newline在等待就医期间，您可以采取以下措施缓解症状：\newline1.保持双手清洁干燥。\newline2.避免使用含有刺激性成分或化学物质的洗涤剂或洗手液清洗双手。\newline3.穿宽松透气的衣服，并尽量避免接触过敏原物质。\newline 4.多饮水，多吃新鲜蔬菜水果以补充维生素和矿物质。\newline5.注意心理调节，减轻压力。\newline(Hello, according to the symptoms you describe, it may be an inflammatory skin reaction caused by eczema. However, in order to determine the cause and treatment plan more precisely, we recommend that you visit a hospital. Your doctor can make a definitive diagnosis by carefully examining your symptoms and asking for information about your history. \newline While waiting to seek medical attention, you can take the following steps to relieve your symptoms: \newline 1. Keep your hands clean and dry. \newline 2. Avoid washing your hands with detergents or hand sanitizers that contain harsh ingredients or chemicals. \newline 3. Wear loose and breathable clothes and try to avoid contact with allergenic substances. \newline4. Drink more water and eat more fresh vegetables and fruits to replenish vitamins and minerals. \newline 5. Pay attention to psychological adjustment and reduce stress.)\\
\midrule
\textbf{HuatuoGPT (w/o RLAIF):} \newline您好，根据您的描述，您可能患有湿疹或者接触性皮肤炎等皮肤病。这种情况通常与皮肤对某些物质过敏有关，例如洗涤剂、化妆品、金属制品等。\newline我建议您先避免使用任何可能会引起过敏反应的物品或环境刺激，如香水、洗洁精、化妆品等等。同时注意保持双手卫生清洁干燥，穿戴柔软透气的手套，避免用手抓挠患处，以免感染加重。\newline如果症状持续存在并无法缓解，您可以到医院就诊，医生会进一步检查确定病因后给予相应的治疗方案。一般情况下，可以采用外用药物进行治疗，比如氢化可的松乳膏（尤卓尔），但具体剂量和使用方法需遵医嘱。\newline(Hello, according to your description, you may be suffering from a skin condition such as eczema or contact dermatitis. This condition is usually associated with skin allergies to certain substances, such as detergents, cosmetics, metal products, etc. \newline I recommend that you first avoid any items or environmental stimuli that may cause allergic reactions, such as perfumes, detergents, cosmetics, etc. Also pay attention to keeping your hands hygienically clean and dry, wear soft and breathable gloves, and avoid scratching the affected area with your hands to avoid aggravating the infection. \newline If the symptoms persist and cannot be relieved, you can go to the hospital, the doctor will further examine to determine the cause of the disease and give the appropriate treatment plan. In general, you can use topical medications for treatment, such as hydrocortisone cream (Euthyrox), but the specific dosage and application methods need to follow medical advice.)\\
\midrule
\textbf{  HuatuoGPT:} \newline您好，根据您的描述，您可能患有汗疱湿疹。这是一种由病毒感染引起的皮肤病，通常会在皮肤摩擦或受热后出现水疱和瘙痒感。\newline为了确诊病情，我需要了解更多信息来确定诊断。请问这些症状出现了多长时间了？是否伴随其他不适感觉，如疼痛、发热或其他皮肤变化等？是否有过敏史或接触过有害物质的历史？\newline如果您没有明显的身体不适，暂时可以尝试以下措施缓解症状：避免穿紧身衣物或使用化学物质的物品，保持皮肤清洁干燥；使用含有天然成分的保湿霜或乳液涂抹患处，以减轻瘙痒和水泡的症状；如果症状严重影响了日常生活，请及时就医咨询医生的建议。\newline(Hello, according to your description, you may have eczema with sweating blisters. This is a skin condition caused by a viral infection that usually results in blisters and itching after rubbing or heat on the skin. \newline In order to diagnose the condition, I need to know more information to be sure of the diagnosis. How long have these symptoms been present? Are they accompanied by other uncomfortable sensations, such as pain, fever or other skin changes? Is there a history of allergies or exposure to harmful substances? \newline If you are not experiencing significant physical discomfort, you can try the following measures to relieve your symptoms for the time being: avoid wearing tight clothing or using chemical substances, keep your skin clean and dry; use a moisturizing cream or lotion with natural ingredients on the affected area to reduce the symptoms of itching and blisters; if your symptoms are severely affecting your daily life, seek medical advice promptly.)\\
\bottomrule
    
\end{tabular}
}
\caption{\label{case study}
Sample responses from four models.
}
\end{table*}


\subsection{Limitation}

We  emphasize the potential risks associated with generation-based medical consultation. The main concern lies in the challenge of verifying the accuracy and correctness of the generated content. In the medical domain, the dissemination of misleading information can have severe ethical implications. Although generative QA has shown promise, especially with the success of models like ChatGPT, they are not yet fully prepared for real-world deployment in the biomedical domain.

While generation methods currently hold great potential, it is important to exercise caution and prudence before deploying them in real-world applications. Further research and development are necessary to refine these models, enhance their accuracy, and establish robust mechanisms for accurateness-checking and error correction. Only through careful scrutiny and continual improvement can we minimize the risks and ethical concerns associated with generation-based medical QA.


\section{LLMs in Medicine}
The language model in the medical field has always been a concern for researchers. The early models were mainly based on the GPT-2 series models to continue pre-training in the domain. BioMedLM\footnote{\url{https://www.mosaicml.com/blog/introducing-pubmed-gpt}} is a domain-specific large language model for biomedicine, trained from 2.7B GPT-2. It is trained on the PubMed Abstracts and PubMed Central portions of the Pile dataset, which contains around 50B tokens and spans a collection of 16 million abstracts and 5 million full-text articles from the biomedical literature. Similarly, BioGPT~\cite{Luo_2022} is a medium GPT-2 model pre-training in medical data collected from the official PubMed website~\footnote{\url{https://ftp.ncbi.nlm.nih.gov/pubmed/}}. For downstream tasks, it uses the soft prompt for fine-tuning training.

Recently, many efforts have attempted to use instruction fine-tuning to enhance the ability for medical consultation on large-scale language models (>6B), as shown in Table 2. \textbf{MEDALPACA}\cite{han2023medalpaca}\footnote{\url{https://github.com/kbressem/medAlpaca}} is a LLaMA model trained on the Medical Meadow, consisting of two
main categories, a collection of established medical NLP tasks reformatted in instruction tuning formats, as well as a crawl of various internet resources. \textbf{ChatDoctor}~\cite{li2023chatdoctor}\footnote{\url{https://github.com/Kent0n-Li/ChatDoctor}} is also a medical LLM trained on the HealthCareMagic-100k dataset based on the LLaMA model. The HealthCareMagic-100k dataset consists of  100k real-world patient-physician
conversations from an online medical consultation site. ChatDoctor has
autonomous knowledge retrieval capabilities by accessing real-time and authoritative
information and answering patient questions based on databases such as Wikipedia to improve the accuracy of the model’s response. \textbf{Baize-healthcare}\footnote{\url{https://huggingface.co/project-baize/baize-healthcare-lora-7B}} is a variant of Baize that is fine-tuned on Medical data (Quora Dialogs and Medical Dialogs). The technique report associated with it has not been published, resulting in limited details being available, as only the model weights were released. \textbf{Visual Med-Alpaca}\footnote{\url{https://github.com/cambridgeltl/visual-med-alpaca}}
is fine-tuned on LLaMA-7B model using a model-generated dataset comprising of manual filtering 54,000 biomedical examples for instruction-tuning purposes, plus the fine-tuned Microsoft GIT model on the Radiology Objects in Context (ROCO) dataset to incorporate visual modality. Recently, \textbf{Med-PaLM2}~\cite{singhal2023expertlevel} was published, which is based on PaLM2 and finetuned in MultiMedQA for Expert-Level Medical Question Answering.

In Chinese, \textbf{DoctorGLM}~\cite{xiong2023doctorglm}\footnote{\url{https://github.com/xionghonglin/DoctorGLM}} is a Chinese Medical LLM trained on Multiple Medical QA datasets based on ChatGLM. It utilizes the training data from ChatDoctor through translation and incorporates Chinese medical dialogues encompassing five departments' QA and MedDialog chat data as part of the training data. \textbf{BenTsao}~\cite{wang2023huatuo}\footnote{\url{https://github.com/SCIR-HI/Huatuo-Llama-Med-Chinese}} is a knowledge-enhanced Chinese Medical LLM trained on over 8K instructions. The instruction is generated from CMeKG~\footnote{\url{https://github.com/SCIR-HI/Huatuo-Llama-Med-Chinese}} by ChatGPT API.\textbf{ MedicalGPT-zh}~\cite{MedicalGPT-zh} is a Chinese medical general model based on ChatGLM-6B LoRA with 16-bit instruction fine-tuning. The dataset for training the model was obtained from Chinese medical knowledge question-and-answer pairs and clinical guideline texts from 28 medical departments.



\begin{table}
\centering
\resizebox{\textwidth}{!}{
\begin{tabular}{llllllllll}
\hline
\textbf{Model Name} & \textbf{Access} & \textbf{Language} & \textbf{Base Model} & \textbf{Training Dataset} & \textbf{Dataset Size} \\
\hline
Med-PaML2 & - & English & PaML2 (340B) & MultiMedQA & 193k  \\
\hline
ChatDoctor & Need request & English & LLaMA (7B)& HealthCareMagic-100K + iCliniq-10K & 110K  \\
\hline
Visual Med-Alpaca & Need request & English & LLaMA (7B) & various medical QA datasets & 54K  \\

\hline
MEDALPACA & Access & English & LLaMA (7B/13B) & Medical Meadow & 160K \\
\hline

Baize-healthcare & Access & English & LLaMA (7B)& Quora Dialogs + Medical Dialogs & 101K \\

\hline
DoctorGLM& Access & Chinese & ChatGLM (6B) & CMD. + MedDialog + ChatDoctor + HearlthcareMagic & 4.4M \\
\hline
BenTsao& Access & Chinese & LLaMA (7B) & CMeKG-8K & 8K  \\

\hline
\end{tabular}
}
\caption{The Popular Medical Language Models Training Data Sources and Model Specifications.}
\label{tab:model-details}
\end{table}






\section{Conclusion}
\label{conclusion}

In conclusion, this paper presents a comprehensive approach to training a reliable and conversational healthcare model by leveraging complementary data sources and incorporating AI model feedback through reinforcement learning. The proposed approach addresses the limitations of relying solely on real or synthetic data and allows for the creation of a model that combines the strengths of both sources. By continuously refining its responses based on feedback, the model can improve its conversational abilities while maintaining the reliability necessary for healthcare applications. Further research in this area holds significant potential for advancing the field of AI in healthcare and improving patient outcomes.

\section*{Acknowledgements}

We thank  Prof. Zhi-Quan Luo and Dr. Ping Li for their support in SRIBD.

{
\small
\bibliographystyle{unsrt}
\bibliography{main}
}

\clearpage

\appendix

\section{Methodology details}
\subsection{Distilled Instructions from ChatGPT}
Following previous work, we use self-instruction to generated the instructions from ChatGPT with the medical seed instructions we manually build and the prompt is shown below:
\label{Prompt_single_distilled}
\begin{verbatim}
你被要求设计20个不同的<角色，指令，输入>三元组，第一行是角色，第二行是该角色希望GPT帮助他提升生活工作效率的指令，第三行是该指令对应的输入。
要求：
1.角色可以很具体，需要跟医疗场景有关，如果是医生的话，甚至可以细化到医疗科室，例如“呼吸内科医生”。
2.每个指令的描述应该是多样化的，指令的类型应该是多样化，动词尽量不要重复，以最大限度地提高多样性。每个指令应该是GPT语言模型能够完成的事情，不能生成绘制图片，不能阅读音频和网页链接; 指令应该是1到2句话的长度，既可以是命令句，也可以是疑问句;指令通常有一个占位符，placeholder，例如“下面这个”或者“某个”，“输入”字段会指定。
3.输入应该为指令的具体例子，提供真实的实质性内容，因为指令可能很空洞，需要用一个具体的输入来限定，输入不能是只有一个链接或者文件名,或者没有特指的“一篇论文”，而应该是具体的内容。输入最好不要超过200字。 
4.角色、指令和输入大多是中文的，角色、指令和输入都不要重复。指令是必须需要，请尽量提供角色和输入。 输入可以为空
20个三元组的清单如下：
\end{verbatim}

The translation is as below:
\begin{verbatim}
You are asked to design 20 different triplets of<characters, instructions, 
input>. The first line is the character, the second line is the instruction 
that the character wants GPT to help him improve his work efficiency, and 
the third line is the corresponding input for the instruction.
requirement:
1. The role can be very specific and needs to be related to the medical scene. 
If it is a doctor, it can even be refined to the medical department, 
such as "respiratory physician".
2. The description of each instruction should be diverse, and the types of 
instructions should be diverse. Verbs should be avoided as much as
possible to maximize diversity. Each instruction should be something 
that the GPT language model can accomplish, unable to generate and 
draw images, unable to read audio and webpage links; Instructions
should be 1-2 sentences in length, which can be either command 
sentences or interrogative sentences; Instructions usually have a 
placeholder, placeholder, such as "this below" or "some", and the 
"input" field will be specified.
3. The input should be a specific example of the instruction, providing 
real substantive content, as the instruction may be empty and need to
be qualified with a specific input. The input should not be just a 
link or file name, or an unspecified 'paper', but rather specific
content. It is recommended to input no more than 200 words.
4. The roles, instructions, and inputs are mostly in Chinese, and the roles, 
instructions, and inputs should not be repeated. Instructions are
mandatory, please provide roles and inputs as much as possible. 
Input can be empty
The list of 20 triples is as follows:
\end{verbatim}

Different from the original self-instruction, we generated role-enhanced instructions and it will be used to generate the output with the following prompt.
\label{Prompt_single_distilled_2}
\begin{verbatim}
假设你是一名经验丰富的[Medical Role]，会对患者给予非常耐心且全面的回答，并且语气温柔亲切，非常受患者喜欢。如果患者没有提供给你足够的信息判断，你会反问他相关问题。而且在诊断最后，你还会给予他一些额外的建议。如果患者提问:
{Question}
那么，你会回答：
\end{verbatim}

The translation is as below:
\label{Prompt_single_distilled_2_en}
\begin{verbatim}
Assuming you are an experienced [Medical Role], you will provide very
patient and comprehensive answers to patients, and your tone will be
gentle and friendly, which is very popular with patients. If the
patient does not provide you with enough information to make a 
judgment, you will ask them relevant questions. And at the end of the 
diagnosis, you will also give him some additional advice. If the 
patient asks: 
{Question}
you will answer:
\end{verbatim}

\subsection{Real-world Instructions from Doctors}
In the experiment, we collect real-world question answering data from web and sample a set of high quality question-answering pairs used for training. Every pair is refined by LLMs. The prompt is shown below:
\label{Prompt_single_real}
\begin{verbatim}
<患者问题>:{Patient_Question}

<回复参考>:{Doctor_Response}

你是HuatuoGPT人工智能模型，基于患者的问题，请你参考回复然后对患者的问题给出回复，说话方式要像医生，并且语气温柔亲切，对患者的询问要回复的更详细更有帮助，在必要时如果无法明确诊断患者的疾病，可以询问患者更多的信息。

<HuatuoGPT回复>:
\end{verbatim}

The translation is as below:
\begin{verbatim}
<Patient Question>:{Patient_Question}

<Response Reference>:{Doctor_Response}

You are HuatuoGPT AI model, based on the patient's question, please refer 
to the response and then give a reply to the patient's question, speak like 
a doctor and have a gentle and kind tone, reply to the patient's query in a
more detailed and helpful way, and ask the patient for more information
if necessary if you cannot clearly diagnose the patient's disease.

<HuatuoGPT Response>:
\end{verbatim}

\subsection{Real-world Conversations with Doctors}
\label{Prompt_conv_real}
We show prompts used for patient LLM and doctor LLM.
Prompt for patient LLM:
\begin{verbatim}
你是一名患者，下面是你的病情，你正在向HuatuoGPT智能医生咨询病情相关的问题，请记住这是一个多轮咨询过程，每次询问要精炼一些，第一次询问要尽可能简单点、内容少一点。
{medical_case}
当你认为整个问诊应该结束的时候请说：再见
\end{verbatim}

Prompt for doctor LLM:
\begin{verbatim}
你是一名经验丰富的医生，会对患者给予非常耐心且全面的回答，说话方式像医生，并且语气温柔亲切，非常受患者喜欢，对患者的询问要回复的更详细更有帮助。如果患者没有提供足够的信息用以诊断，你要反问他相关问题来获取更多信息来做出诊断，做出诊断后你还会给予他一些额外详细的建议。注意，你只能接收患者的描述没法看到图片之类的材料或附件。
如果无法做出明确的诊断，请询问出患者更多的病情信息，最后给出的诊断结果可以是：
{doctor_diagnosis}
\end{verbatim}

\subsection{Prompt for AI feedback}
\begin{verbatim}
Here is a conversation history:
[History]
{history}
[End of History]
Here is the final question and the standard answer:
[Question]
{query}
[End of question]
[Standard answer]
{doctor_response}
[End of standard answer]
Based on the conversation history, user question, and standard answer,
please rate the following two AI responses on a scale of 1 to 10,
considering accuracy, conciseness, and similarity to the standard answer.
Please provide the ratings in the following format: "Rating A: [score];
Rating B: [score]".
[Assistant A]
{Response_A}
[End of Assistant A]
[Assistant B]
{Response_B}
[End of Assistant B]
\end{verbatim}

\end{CJK}
\end{document}